\documentclass{article}

\usepackage{microtype}
\usepackage{graphicx}
\usepackage{subfigure}
\usepackage{booktabs} 

\usepackage{hyperref}

\usepackage[accepted]{icml2024}

\usepackage{amsmath}
\usepackage{amssymb}
\usepackage{mathtools}
\usepackage{amsthm}
\usepackage{kotex}
\usepackage{bbm}
\usepackage{enumitem}
\usepackage{algorithm}
\usepackage{algorithmic}
\usepackage{times}
\usepackage{epsfig}
\usepackage{graphicx}
\usepackage{color}
\usepackage{multirow}
\usepackage{colortbl}
\usepackage{xcolor}
\usepackage{oplotsymbl}
\usepackage{MnSymbol}
\usepackage{bbding}
\usepackage{mwe} 
\usepackage{wrapfig}
\DeclareMathOperator*{\argmax}{argmax}

\newcommand{\ie}{\textit{i}.\textit{e}.}

\usepackage[capitalize,noabbrev]{cleveref}

\theoremstyle{plain}

\theoremstyle{definition}

\theoremstyle{remark}

\usepackage[textsize=tiny]{todonotes}

\icmltitlerunning{Mitigating the Bias in the Model for Continual Test-Time Adaptation
}

\begin{document}

\twocolumn[
\icmltitle{Mitigating the Bias in the Model for Continual Test-Time Adaptation}




\begin{icmlauthorlist}
\icmlauthor{Inseop Chung}{snu}
\icmlauthor{Kyomin Hwang}{snu}
\icmlauthor{Jayeon Yoo}{snu}
\icmlauthor{Nojun Kwak}{snu}
\end{icmlauthorlist}

\icmlaffiliation{snu}{Graduate School of Convergence Science and Technology, Seoul National University, Seoul, South Korea}

\icmlcorrespondingauthor{Inseop Chung}{jis3613@snu.ac.kr}
\icmlcorrespondingauthor{Nojun Kwak}{nojunk@snu.ac.kr}

\icmlkeywords{Test-Time Adaptation, Continual Test-Time Adaptation, Domain Adaptation}

\vskip 0.3in
]



\printAffiliationsAndNotice{\icmlEqualContribution} 

\begin{abstract}
Continual Test-Time Adaptation (CTA) is a challenging task that aims to adapt a source pre-trained model to continually changing target domains. In the CTA setting, a model does not know when the target domain changes, thus facing a drastic change in the distribution of streaming inputs during the test-time. The key challenge is to keep adapting the model to the continually changing target domains in an online manner. We find that a model shows highly biased predictions as it constantly adapts to the chaining distribution of the target data. It predicts certain classes more often than other classes, making inaccurate over-confident predictions. This paper mitigates this issue to improve performance in the CTA scenario. 
To alleviate the bias issue, we make class-wise exponential moving average target prototypes with reliable target samples and exploit them to cluster the target features class-wisely. Moreover, we aim to align the target distributions to the source distribution by anchoring the target feature to its corresponding source prototype. 
With extensive experiments, our proposed method achieves noteworthy performance gain when applied on top of existing CTA methods without substantial adaptation time overhead.
\end{abstract}

\section{Introduction}\label{sec:intro}

Data \emph{distribution shifts} is a problem which the distribution of data given at test-time is different from that of the training data. This is because the DNNs heavily rely on the assumption that test-time data are independent and identically distributed (i.i.d.) with the training data which is very unlikely in real-world scenarios~\citep{hendrycks2019robustness, koh2021wilds}.
Test-time adaptation (TTA)~\citep{sun2020ttt, wang2020tent, zhang2022memo} resolves this issue by adapting the model to the target data given at test-time. Since the target data are unlabeled, the adaptation is done in an unsupervised and online manner which means that the model has to predict and adapt immediately upon the arrival of the test samples.
TTA generally assumes that the access to the source data during test-time is infeasible due to privacy/storage concerns and legal constraints, hence the only available during the test-time is the access to the target data and the off-the-self source pre-trained model.
Recently, another line of research in TTA called continual test-time adaptation (CTA)~\citep{wang2022continual, niu2022eata} is introduced. Different from the conventional TTA setting which assumes adapting a model to a single fixed stationary target distribution, CTA assumes the target distribution changes over time. The timing of the distribution changes is not provided.
Therefore, the model needs to constantly adapt to shifting target data distributions, and it is not feasible to reset the model to its initial source pre-trained weights when distribution changes occur. This makes CTA an extremely challenging task resembling the real-world scenarios where the input distribution may change continually and abruptly without prior notice (e.g. entering a tunnel during autonomous driving). 

\begin{figure}[t]
    \centering
    \includegraphics[width = 1.0\linewidth]{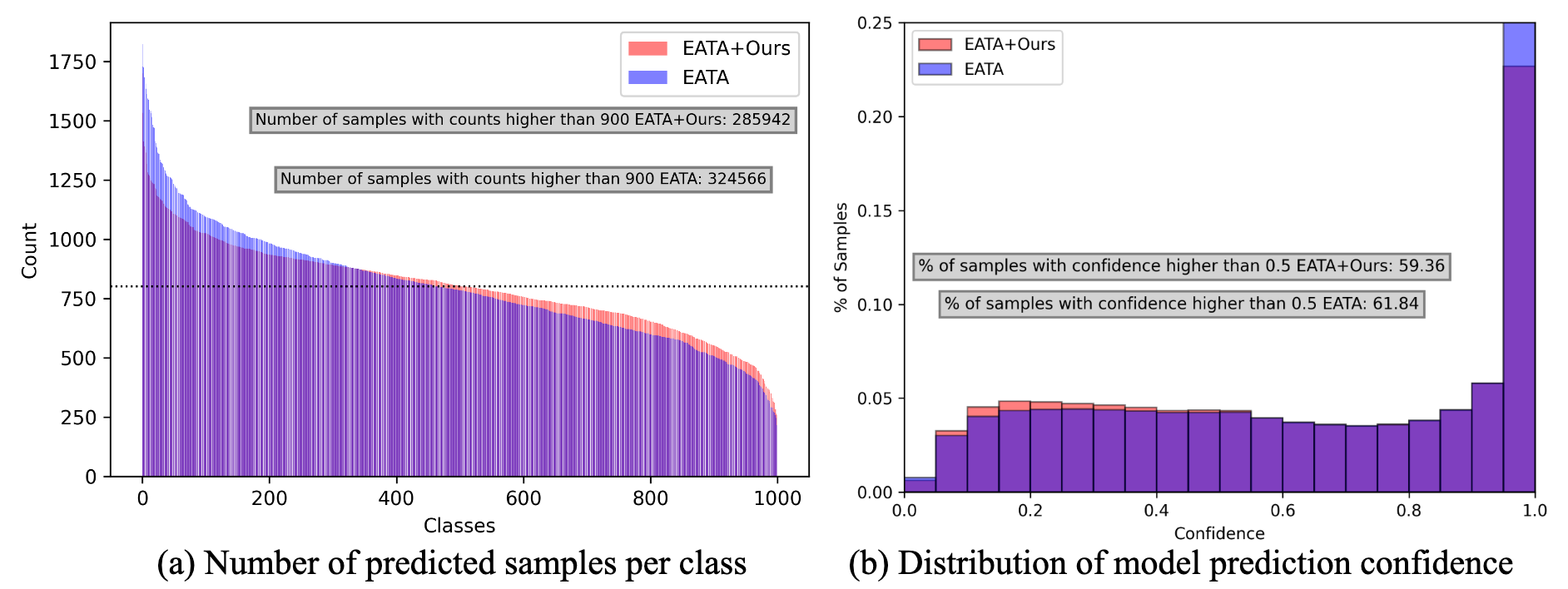}
    \caption{Comparison of the number of predicted samples per class and distribution of confidence between EATA and EATA+\textbf{Ours}.}
    \label{fig:bias_in_model}
\end{figure}

Due to its intricate nature, the model is susceptible to confirmation bias~\citep{arazo2020pseudo}, where it tends to overfit to the incoming target data while continuously adapting in an online manner. We observe this results in highly biased and mis-calibrated model predictions. Fig.~\ref{fig:bias_in_model} shows the number of predicted samples per class and the distribution of prediction confidence of the model trained by EATA \citep{niu2022eata}, one of the state-of-the-art CTA algorithm, and EATA+\textbf{Ours} using the ImageNet-C~\citep{deng2009imagenet} benchmark. The horizontal dotted line in Fig.~\ref{fig:bias_in_model} (a) indicates the actual number of samples assigned to each class. The classes are sorted in descending order of the number of predicted samples for clarity. Even though EATA shows decent average accuracy in ImageNet-C (49.81\%), its prediction is highly biased to favor certain classes more often while avoiding predictions for others. Also, Fig.~\ref{fig:bias_in_model} (b) shows that EATA makes 25\% of its prediction with confidence higher than 0.95, highlighting a significant issue of overconfidence in the model.

To overcome the aforementioned bias in the model and to further improve its performance in CTA scenario, this paper presents a pair of straightforward yet highly effective techniques: the exponential moving average (EMA) target domain prototypical loss and source distribution alignment via prototype matching. 
The EMA prototypical loss maintains a prototype for each class by continuously updating each prototype with the features of reliable target samples given at test-time in an EMA fashion. These EMA target prototypes are utilized to organize the target features into distinct classes by pulling them closer to their corresponding EMA prototypes while simultaneously pushing them away from other irrelevant prototypes. The EMA prototypical loss effectively captures the changing target distribution and leverages it for class-specific clustering. Its goal is to prevent an undue bias towards current target distributions and, instead, adeptly capture and adapt to changing target distributions, thereby mitigating the bias issue.
On the other hand, to prevent the model from drifting too far away from the pre-trained source distribution, we align the target data distribution to the source distribution by minimizing the distance between the target feature and its corresponding source prototype. Aligning the distribution between source and target is a common strategy in domain adaptation~\citep{tzeng2017adversarial, long2018conditional} which has also been employed in TTA method~\citep{su2022revisiting}. Nonetheless, it relies on the strong assumption that both domains follow the Gaussian distribution and employ complex distance metric such as KL-Divergence. In contrast, our method takes a simpler approach: we directly minimize the mean squared error distance between each target feature and its corresponding source prototype.
As depicted in Fig.~\ref{fig:bias_in_model}, our introduced terms effectively alleviate the bias in predictions. EATA+\textbf{Ours} exhibits reduced inclination to favor specific classes, resulting in a more balanced distribution of predictions across classes compared to EATA. The overconfident predictions is also mitigated along with improved average accuracy (51.32\%). Contributions of this paper are as follows:

\begin{itemize}
  \item The proposed method is seamlessly applicable to existing approaches without additional parameters or requiring access to the source domain data at test-time which transforms it into a simple plug-and-play component. 
  
  \item Through comprehensive experiments on ImageNet-C and CIFAR100-C, the proposed method is shown to be compatible with other CTA methods and able to substantially improve the accuracy without significant adaptation time overhead.
  
  \item We conduct an in-depth analysis of our proposed method, highlighting its capability to mitigate the bias of the model by restraining from making over-confident predictions and fostering more calibrated confidence.
  
\end{itemize}

\begin{figure*}[t]
    \centering
    \includegraphics[width = 0.9\linewidth]{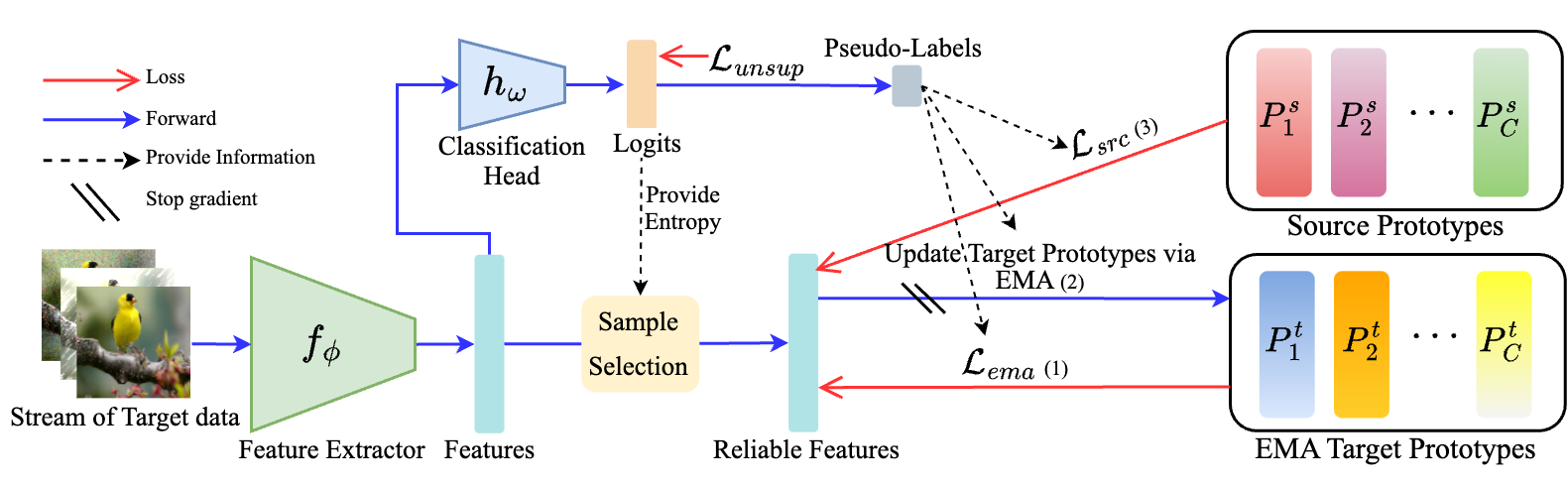}
    \caption{
    Before deploying the model, we generate the source prototypes ($P^s$s) using the subset of source data and the source pre-trained feature extractor, $f_{\phi_0}$. After the model is deployed to the target domain, the model adapts to the target data by minimizing our proposed terms $\mathcal{L}_{ema}$ and $\mathcal{L}_{src}$ along with $\mathcal{L}_{unsup}$. We construct class-wise target prototypes ($P^t$s) that are updated with target features via EMA manner. We utilize both $P^t$s and $P^s$s to compute $\mathcal{L}_{ema}$ and $\mathcal{L}_{src}$ respectively. Note that $\mathcal{L}_{ema}$ is first computed and then followed by updating the $P^t$s subsequently. The dotted line indicates providing required information such as entropy and pseudo-label of input.
    }
    \label{fig:overall}
\end{figure*}

\section{Related Works}\label{related_works}
\subsection{Test-Time Adaptation}
Recently, test-time adaptation (TTA) has garnered substantial attention, adapting models to specific test domains during inference-time after being deployed to the target data. TTA shares similarities with source-free domain adaptation (SFDA)~\citep{liang2020shot}, in the aspect of adapting the off-the-shelf source pre-trained model to the target domain without accessing source data. However, TTA differs from SFDA in that it is an online learning approach relying solely on the incoming target samples given at test-time without repetitively accessing a large amount of unlabeled target domain data. This feature makes TTA more challenging in that overall information such as knowing the target domain distribution~\citep{sun2016dcoral} or clustering the target features~\citep{liang2020shot} is not available.
Many studies~\citep{wang2020tent, niu2022eata, lim2023ttn} efficiently adapt models to the test domain by updating only the batch normalization layer, following the research~\citep{bn2020schneider} that only replacing the statistics for batch normalization without learning can effectively address domain shifts. These methods~\citep{wang2020tent, niu2022eata} adapt the model to the target domain via entropy minimization loss to make the predictions more confident. Alternatively, there are approaches~\citep{su2022revisiting, jung2022cafa} that update the entire backbone so that the distribution of the target domain feature has similar statistics to that of the source on the premise that the statistics of the source domain features are known. Some other methods~\cite{iwasawa2021t3a,t3adapter2023jang} entirely freeze the backbone and solely modify the classifier by leveraging prototypes derived from target domain features based on pseudo-labels. Additionally, some methods~\citep{sun2020ttt, ttaps2022bartler} modify the model architecture during source domain training to incorporate self-supervised losses for the target data during test-time.

\subsection{Continual Test-Time Adaptation (CTA)}
In practice, the distribution of the test domain can exhibit continuous changes or have correlations among continuously incoming samples, whereas TTA relies on a strong assumption that test-time data follow i.i.d, meaning that the distribution of the test-time data does not change and stays stationary. CoTTA~\citep{wang2022continual} first suggests the problem of continual test-time adaptation and proposes the corresponding problem setting. It identifies the problem of error accumulation in existing TTA methods when the distribution of test-time data changes and addresses it by introducing a teacher-student framework and ensuring various augmented test samples to have consistent predictions along with stochastic restoration of the weights. Following this, \citet{brahma2023probabilistic} and \citet{dobler2023robust} also utilize a teacher-student structure, employing regularization based on the importance of weights and using symmetric cross-entropy loss, respectively. Additionally, \citet{niu2022eata}, which considers the confidence and diversity of samples for model updates, has proven to be effective in the context of CTA. Building upon existing TTA methods, \citet{song2023ecotta, hong2022mecta} have proposed techniques to diminish memory consumption, thereby promoting efficient adaptation in CTA.

\subsection{CTA under Dynamic Scenarios}
Recently, there has been many attempts to consider dynamic scenarios in CTA~\cite{gong2022note, niu2023towards, yuan2023robust, gong2023sotta}. NOTE~\cite{gong2022note} and RoTTA~\cite{yuan2023robust} point out that real-world data are often temporally correlated (non-i.i.d) and propose robust CTA methods against non-i.i.d. test data. SAR~\cite{niu2023towards} considers test data with mixed domains shifts, single sample batch and imbalanced label shift. Recently, SoTTA~\cite{gong2023sotta} claims that, in real-world settings, extraneous samples outside the model's scope, such as unseen objects, noise, and adversarial samples created by malicious users, can be provided as inputs and proposes a way to screen out these noisy samples during CTA.

\section{Problem Definition}
Given a model, $g_{\theta_0}$, pre-trained on a source domain $D^s = \{x^s_n, y^s_n\}^{N^s}_{n=1}$, CTA is a task of adapting $g_{\theta_0}$ to the unlabeled target data which its domain continually changes, $D^k = \{x^k_m\}^{N^k}_{m=1}$ ($k$ refers to the target domain index) with an unsupervised objective, $\mathcal{L}_{unsup}$. The target domain data arrive sequentially and their domain changes over time ($k = 1, \dots, K$). The model only has access to the data of the current time step and has to predict and adapt instantly upon the arrival of the inputs for future steps, \ie, $\theta_t \rightarrow \theta_{t+1}$. 
As mentioned earlier, the model is not aware of when the target domain changes, so it has to deal with suddenly changing input distribution. $\mathcal{L}_{unsup}$ can take the form of entropy minimization loss which is used to optimize only the affine parameters of batch normalization layer~\citep{wang2020tent, niu2022eata} or consistency loss to optimize the whole parameters~\citep{wang2022continual, dobler2023robust}. The evaluation of the model is determined by test-time predictions in an online manner. 

\section{Proposed Method}
\subsection{EMA Target Domain Prototypical Loss}
EMA target prototypical loss comprises two distinct steps, one is categorizing the features of target inputs by classes utilizing the EMA target prototypes and the other is updating the prototypes with features of reliable target samples in an exponential moving average manner.
A classification model, $g_{\theta}$, consists of a feature extractor $f_{\phi}$ and a classification head $h_{\omega}$. Each weight vector $\omega_c \in \mathbb{R}^d$ in $\omega \in \mathbb{R}^{C \times d}$ can be considered as the template for class $c$ where $C$ is the number of classes and $d$ is the dimension of the extracted feature, $f_{\phi}(x) \in \mathbb{R}^{d}$. Therefore, we initialize the EMA target prototypes as the weights of $h$, hence $P^t_c = \frac{\omega_c}{||\omega_c||_2}$. $P^t_c$ and $\omega_c$ refer to the EMA target prototype and the head weight of class $c$, respectively. 
We normalize $\omega_c$ to eliminate the difference in magnitudes between $\omega_c$ and the extracted target feature $f_{\phi}(x^t)$ when updating the target prototypes via~(\ref{eq:ema_update}). 
There are $C$ number of EMA target prototypes, which we utilize to categorize the streaming target inputs into classes. This is achieved by minimizing the cross-entropy loss using the pseudo-labels.
\begin{algorithm}[t]
    \caption{The pseudo code of our proposed CTA process for $K$ number of target domains.}
    \label{alg:overall}
    \begin{algorithmic}[1]
    \REQUIRE{$K$ number of target domains $\{D^k = \{x^k_m\}^{N^k}_{m=1}\}_{k=1}^{K}$, the source pre-trained model $g_{\theta_0}(\cdot)$, Source sub-samples $D^s\small{=}\{x^s_n\}_{n=1}^{N^s}$, batch size $B$.}
    \STATE Generate the source prototype for each class, $P^s_{c} = \frac{1}{N^s_c}\sum_{i=1}^{N^s_c} f_{\phi_0}(x^s_i)$.
    \STATE Initialize each EMA target prototype, $P^t_{c}$ as $\frac{\omega_c}{||\omega_c||_2}$
    \FOR{a domain $k$ in $K$}
        \FOR{a batch $\textbf{x}\small{=}\{x^k_b\}_{b=1}^{B}$ in $D^k$}
        \STATE Forward the batch and make predictions, $\textbf{z}=g_{\theta}(\textbf{x})$
        \STATE Compute $\mathcal{L}_{unsup}$
        \STATE Identify reliable inputs with low entropy
        \STATE Compute $\mathcal{L}_{ema}$ and $\mathcal{L}_{src}$ only with the features of reliable target inputs.
        \STATE Update $P^t$s via (\ref{eq:ema_update})
        \STATE Optimize model by minimizing $\mathcal{L}_{overall}$.
        \ENDFOR
    \ENDFOR
    \end{algorithmic}
\end{algorithm}
However, before computing the loss, we first identify reliable target samples as proposed in \citep{niu2022eata}, which excludes samples with high entropy, thus low confidence. Given a batch of target data, $\textbf{x}^t \in \mathbb{R}^{B \times C \times H \times W}$, for each sample $x^t$ in $\textbf{x}^t$, we calculate its entropy estimated by the model $g_{\theta}$, $H_{\theta}(x^t)$.
Then, we filter out samples with entropy higher than the pre-defined entropy threshold, $E_0$. The remaining samples are the reliable samples with low-entropy denoted as $\tilde{\textbf{x}}^t$. For each sample $\tilde{x}^t$ in $\tilde{\textbf{x}}^t$, we obtain its pseudo-label $\tilde{y}^t = \argmax_c g_{\theta}(\tilde{x}^t)_c$ and compute the following loss:

\begin{align}\label{eq:pseudo_ce}
    \mathcal{L}_{ema}=- \log(\frac{\exp(f_{\phi}(\tilde{x}^t) \cdot \frac{P^t_{\tilde{y}^t}}{||P^t_{\tilde{y}^t}||_2})}{\sum_c^C \exp(f_{\phi}(\tilde{x}^t) \cdot \frac{P^t_{c}}{||P^t_{c}||_2})}).
\end{align}
We dot-product $f_{\phi}(\tilde{x}^t)$ with every EMA target prototype $P^t_{c}$ and apply softmax operation, then maximize its similarity with the target prototype of the pseudo-label, $P^t_{\tilde{y}^t}$, by minimizing $\mathcal{L}_{ema}$. $\mathcal{L}_{ema}$ assures $f_{\phi}(\tilde{x}^t)$ to have high similarity with $P^t_{\tilde{y}^t}$ and low similarity with other remaining $P^t$s. $\mathcal{L}_{ema}$ is designed to back-propagate only to the $f_{\phi}$ and not to the $P^t$s.
Upon computing $\mathcal{L}_{ema}$, we proceed to update $P^t$s in an EMA manner using the features of reliable samples and their pseudo-labels as outlined below:

\begin{align}\label{eq:ema_update}
    P^t_{\tilde{y}^t} = \alpha \cdot P^t_{\tilde{y}^t} \plus (1-\alpha) \cdot \frac{f_{\phi}(\tilde{x}^t)}{||f_{\phi}(\tilde{x}^t)||_2}.
\end{align}

Here, $\alpha$ is the blending factor. We normalize the target feature ($\frac{f_{\phi}(\tilde{x}^t)}{||f_{\phi}(\tilde{x}^t)||_2}$) as we normalized $\omega_c$ when initializing $P^t_c$. We detach $f_{\phi}(\tilde{x}^t)$ in order to stop gradient signal to $f_{\phi}$.
If there exists $N_{c}$ number of samples with the same pseudo-label in a batch, we use the average of their features ($\frac{1}{N_{c}}\sum_{i=1}^{N_{c}} f_{\phi}(\tilde{x}^t_i)$) for updating the target prototype, $P^t_c$. 
As new batches of target data steam in, $P^t$s are updated with features of new incoming target data in an EMA fashion. 
The individual magnitudes of each $P^t$ can vary, potentially leading to inaccuracies in the results. To address this issue and ensure consistency in magnitudes, we normalize each $P^t_c$ before performing the dot product with $f_{\phi}(\tilde{x}^t)$ as described in~(\ref{eq:pseudo_ce}).
Please note that $\mathcal{L}_{ema}$ is computed first and then followed by the update of $P^t$ using (\ref{eq:ema_update}) with $f_{\phi}(\tilde{x}^t)$, not the other way around. Also, it is important to mention that $P^t$s are not employed to classify the target input for model evaluation but solely for calculating the loss $\mathcal{L}_{ema}$. The model evaluation is measured by $z=g_{\theta}(x^t)$, with the head of the model, $h$. It is different from T3A~\citep{iwasawa2021t3a} which builds an actual classifier for evaluation with features of target samples given at test-time.

In short, (\ref{eq:pseudo_ce}) organizes the target feature into separate classes by enhancing its similarity with the corresponding EMA target prototype while (\ref{eq:ema_update}) updates class-specific prototypes with the target data features in an EMA manner to gradually reflect the changing target distribution. The purpose is to mitigate the bias in the model by preventing it from being ovetfitted to the current target data but rather to capture more general target distribution than can handle the changing target distribution.

\subsection{Source Distribution Alignment via Prototype Matching}
Prior to deploying the model to the target domain for testing, we generate the source prototype for each class in advance using the subset of the source domain data and the the source pre-trained feature extractor $f_{\phi_0}$. More precisely, we sample a maximum of 100,000 data from the source train set. A source prototype for class $c$ is computed as an average of features extracted by $f_{\phi_0}$, hence $P^s_{c} = \frac{1}{N^s_c}\sum_{i=1}^{N^s_c} f_{\phi_0}(x^s_i)$, where $N^s_c$ is the number of samples with class label $c$ in the subset. There exists $C$ number of source prototypes generated before test-time and are saved in memory to be used later at the test-time adaptation phase. During the test-time, we minimize the mean squared error (MSE) distance between the target feature and the source prototype corresponding to the pseudo-label of the target feature.  
\begin{align}\label{eq:src_alignment}
    \mathcal{L}_{src}=||P^s_{\tilde{y}^t} - f_{\phi}(\tilde{x}^t)||^2_2.
\end{align}
Similar to EMA target prototypical loss, we calculate the above source distribution alignment loss only with the reliable samples, $\tilde{\textbf{x}}^t$.
The intention of $\mathcal{L}_{src}$ is to restrain the model from deviating excessively from the pre-trained source distribution and to align the distributions of the target and the source data, thereby mitigating the impact of distribution shift.

\subsection{Overall Objective}
The overall objective of our proposed continual test-time adaptation method is as follows :
\begin{align}\label{eq:overall}
    \mathcal{L}_{overall}=\mathcal{L}_{unsup} + \lambda_{ema}\mathcal{L}_{ema} + \lambda_{src}\mathcal{L}_{src}
\end{align}
$\mathcal{L}_{unsup}$ represents the unsupervised loss employed in the particular method to which our proposed approach is being applied. Our suggested loss components, $\mathcal{L}_{ema}$ and $\mathcal{L}_{src}$, can be integrated into existing methods with respective trade-off terms, $\lambda_{ema}$ and $\lambda_{src}$. Alternatively, they can be employed independently as well, without the inclusion of $\mathcal{L}_{unsup}$. Fig.~\ref{fig:overall} illustrates the overall process of our proposed method and the pseudo code of our proposed CTA scheme is summarized in Alg.~\ref{alg:overall}.

\section{Experiments}\label{sec:exp}

\textbf{Datasets and models.} We evaluate our proposed method on two widely used test-time adaptation benchmarks, ImageNet-C~\citep{deng2009imagenet} and CIFAR100-C~\citep{krizhevsky2009learning}. Both datasets corrupts the test set of the original dataset with 15 different kinds of corruptions with 5 different levels of severity from four different categories (noise, blur, weather, digital)~\citep{hendrycks2019robustness}. We conduct experiments with the highest level 5. Other than these 15 corrupted target domains, we also perform test-time adaptation on the original clean test set as the last domain to validate how the model has preserved performance on the source domain. We employ ResNeXt29-32$\times$4d pre-trained by AugMix~\citep{hendrycks2019augmix} and ResNet50 pre-trained by \citep{hendrycks2021many} as the source pre-trained models for CIFAR100-C and ImageNet-C, respectively. Both models are trained on the original training set of CIFAR-100 and ImageNet.

\textbf{Evaluation.}
The model is initialized as the source pre-trained weights before test-time adaptation. As the test-time adaptation initiates, batches of target data stream into the model sequentially for prediction and adaptation. The target domain changes when the model encounters all samples of the current target domain, but the domain change information is not given to the model. We report the average classification accuracy of 3 runs for each domain.

\textbf{Implementation Details.} 
Since our proposed method is compatible with existing methods, we adhere to the implementation details of each method to which our approach is applied, including the choice of optimizer and hyper-parameters. 
To ensure a fair comparison, we conduct all experiments using a consistent batch size of 64 across all methods. The entropy threshold, $E_0$ is set to $0.4 \times \ln{C}$ following \citep{niu2022eata}.
$\alpha$, $\lambda_{ema}$ and $\lambda_{src}$ are empirically set to 0.996, 2.0 and 50 when applied on existing method.
However, when our proposed method is employed independently without integration into existing methods, $\lambda_{src}$ is set to 20. and we use SGD with a learning rate of 0.00025, momentum of 0.9 and update only the batch normalization layers as done in previous works~\citep{wang2020tent, niu2022eata}.
More implementation details are in appendix~\ref{app:imp_details}.

\subsection{Performance Comparison} \label{sec:performance_comparison}

\begin{table}[t]
\centering
\caption{Classification accuracy (\%) for the comparison of CTA performance on ImageNet-C using the highest corruption level 5.}\label{tab:IN}
\scalebox{0.4}{
\tabcolsep4pt
\begin{tabular}{l|llllllllllllllll|c}\hline
Time & \multicolumn{16}{l|}{$t\xrightarrow{\hspace*{16.5cm}}$}& \\ \hline
Method & \rotatebox[origin=c]{70}{Gauss.} & \rotatebox[origin=c]{70}{shot} & \rotatebox[origin=c]{70}{impulse} & \rotatebox[origin=c]{70}{defocus} & \rotatebox[origin=c]{70}{glass} & \rotatebox[origin=c]{70}{motion} & \rotatebox[origin=c]{70}{zoom} & \rotatebox[origin=c]{70}{snow} & \rotatebox[origin=c]{70}{frost} & \rotatebox[origin=c]{70}{fog}  & \rotatebox[origin=c]{70}{bright.} & \rotatebox[origin=c]{70}{contrast} & \rotatebox[origin=c]{70}{elastic.} & \rotatebox[origin=c]{70}{pixelate} & \rotatebox[origin=c]{70}{jpeg} & \rotatebox[origin=c]{70}{original} & Mean \\ \hline
Source & 2.21 & 2.93 & 1.85 & 17.92 & 9.82 & 14.79 & 22.50 & 16.88 & 23.31 & 24.42 & 58.94 & 5.44 & 16.96 & 20.61 & 31.65 & 76.13 & 21.65 \\ \hline
T3A & 15.03 & 15.61 & 16.09 & 16.05 & 16.16 & 17.79 & 20.66 & 22.32 & 23.48 & 25.81 & 29.12 & 28.16 & 29.32 & 30.62 & 31.23 & 33.71 & 23.20 \\
TTAC & 23.47 & 32.33 & 32.88 & 24.52 & 29.82 & 40.00 & 47.73 & 42.58 & 40.00 & 50.16 & 61.72 & 26.64 & 47.73 & 51.43 & 45.27 & 66.49 & 41.42 \\
TSD & 15.23 & 15.78 & 15.78 & 15.06 & 15.29 & 26.29 & 38.81 & 34.35 & 33.14 & 47.89 & 65.16 & 16.83 & 44.03 & 48.82 & 39.82 & 75.15 & 34.21 \\
SAR & 30.23& 37.72& 37.18& 27.13& 29.55& 34.52& 41.75& 35.80& 35.33& 46.13& 57.85& 31.20& 46.08& 49.53& 46.17& 64.63& 40.67 \\
RoTTA & 17.05 &23.42 &25.30 &21.48 &19.50 &18.87 &22.39 &21.31 &22.02 &23.61 &39.43 &14.84 &26.72 &25.04 &25.58 &39.83 &24.15 \\
\textbf{Ours}-Only & 32.88 &40.98 &39.78 &29.84 &32.18 &39.04 &45.79 &42.35 &41.54 &52.42 &63.15 &43.74 &52.51 &56.88 &52.86 &69.39 &\textbf{45.96}\\ \hline
TENT & 24.69 & 32.81 & 32.72 & 24.28 & 26.03 & 30.29 & 37.89 & 30.40 & 28.46 & 36.51 & 49.58 & 18.16 & 32.99 & 35.68 & 30.60 & 49.94 & 32.56 \\
TENT + \textbf{Ours} & 30.93 &39.67 &39.24 &29.85 &32.26 &39.28 &45.99 &41.85 &40.57 &50.80 &62.24 &41.84 &49.68 &53.14 &47.55 &62.81 &\textbf{44.23} \\\hline
EATA & 34.66 & 40.40 & 39.39 & 34.08 & 34.99 & 46.51 & 52.82 & 50.33 & 45.83 & 59.12 & 67.27 & 45.17 & 57.13 & 59.99 & 55.46 & 73.80 & 49.81 \\
EATA + TTAC & 35.64 & 41.44 & 40.57 & 35.59 & 37.14 & 48.67 & 54.56 & 51.69 & 46.73 & 60.34 & 67.98 & 46.58 & 58.04 & 61.22 & 56.18 & 74.40 & 51.05\\ 
EATA + \textbf{Ours} & 36.17 & 41.77 & 40.83 & 35.98 & 37.24 & 48.89 & 54.28 & 52.15 & 47.46 & 60.23 & 67.94 & 48.01 & 58.26 & 61.26 & 56.37 & 74.20 & \textbf{51.32} \\ \hline
CoTTA & 16.15 & 18.53 & 19.91 & 18.52 & 19.58 & 31.13 & 43.07 & 36.92 & 36.15 & 51.18 & 65.35 & 23.50 & 47.71 & 52.17 & 44.82 & 73.99 & 37.42\\
CoTTA + \textbf{Ours} & 30.06 & 37.51 & 36.72 & 26.86 & 30.65 & 42.34 & 49.64 & 47.53 & 44.15 & 56.65 & 67.13 & 37.73 & 55.98 & 59.81 & 54.68 & 73.17 & \textbf{46.91} \\ \hline
RMT & 28.45 & 36.07 & 36.39 & 29.83 &29.00 & 35.22 & 39.58 & 40.04 & 36.08 & 49.35 & 54.02 & 36.67 & 48.62 & 52.28 & 48.65 & 66.63 & 41.68 \\
RMT + \textbf{Ours} & 29.60 & 37.85 & 38.26 & 31.60 & 30.98 & 36.46 & 40.56 & 42.06 & 38.24 & 46.31 & 54.19 & 38.02 & 50.73 & 53.24 & 51.24 & 65.14 & \textbf{42.78} \\ \hline
\end{tabular}
}
\end{table}

\begin{table}[t]
\centering
\caption{Classification accuracy (\%) for the comparison of CTA performance on CIFAR100-C using the highest corruption level 5.}\label{tab:cifar100}
\scalebox{0.4}{
\tabcolsep4pt
\begin{tabular}{l|llllllllllllllll|c}\hline
Time & \multicolumn{16}{l|}{$t\xrightarrow{\hspace*{16.5cm}}$}& \\ \hline
Method & \rotatebox[origin=c]{70}{Gauss.} & \rotatebox[origin=c]{70}{shot} & \rotatebox[origin=c]{70}{impulse} & \rotatebox[origin=c]{70}{defocus} & \rotatebox[origin=c]{70}{glass} & \rotatebox[origin=c]{70}{motion} & \rotatebox[origin=c]{70}{zoom} & \rotatebox[origin=c]{70}{snow} & \rotatebox[origin=c]{70}{frost} & \rotatebox[origin=c]{70}{fog}  & \rotatebox[origin=c]{70}{bright.} & \rotatebox[origin=c]{70}{contrast} & \rotatebox[origin=c]{70}{elastic.} & \rotatebox[origin=c]{70}{pixelate} & \rotatebox[origin=c]{70}{jpeg} & \rotatebox[origin=c]{70}{original} & Mean \\ \hline
Source & 27.02 & 32.00 & 60.64 & 70.64 & 45.91 & 69.19 & 71.21 & 60.53 & 54.18 & 49.70 & 70.48 & 44.91 & 62.79 & 25.29 & 58.77 & 78.90 & 55.14 \\ \hline
T3A & 28.10 & 36.47 & 59.70 & 67.25 & 43.91 & 67.07 & 69.93 & 57.42 & 50.83 & 45.34 & 69.55 & 44.13 & 58.64 & 23.52 & 55.77 & 76.82 & 53.40 \\
TTAC & 58.86 & 63.63 & 61.46 & 72.89 & 59.45 & 70.86 & 72.74 & 65.13 & 66.56 & 59.76 & 73.24 & 68.46 & 63.48 & 67.28 & 60.36 & 75.83 & 66.25 \\ 
TSD & 56.87 & 58.64 & 56.23 & 71.62 & 57.45 & 69.56 & 71.31 & 64.22 & 64.44 & 57.38 & 72.88 & 68.83 & 63.41 & 66.06 & 58.07 & 75.32 & 64.52 \\
SAR & 59.03& 63.80& 62.28& 73.45& 61.81& 71.32& 73.76& 67.38& 68.78& 63.19& 74.28& 71.40& 67.27& 70.18& 62.19& 76.61& 67.92 \\
RoTTA & 51.65 &54.96 &54.57 &70.15 &57.95 &70.93 &73.91 &68.38 &69.38 &62.91 &75.20 &71.08 &67.60 &70.65 &63.50 &76.74 &66.22 \\
\textbf{Ours}-Only & 60.62 &66.08 &64.45 &73.79 &62.52 &71.79 &74.23 &67.98 &69.29 &65.34 &73.91 &72.15 &67.04 &70.55 &62.09 &75.66 &\textbf{68.59}\\ \hline
TENT & 58.13 & 62.58 & 61.43 & 73.82 & 61.24 & 71.67 & 73.73 & 67.09 & 68.39 & 61.85 & 74.80 & 71.27 & 66.98 & 70.13 & 61.51 & 77.13 & 67.61 \\
TENT + \textbf{Ours} & 60.24 & 65.56 &63.48 &73.96 &62.64 &72.16 &74.67 &68.24 &69.67 &64.72 &74.66 &73.08 &67.41 &71.01 &62.48 &77.12 &\textbf{68.82} \\ \hline
EATA & 59.91 & 63.92 & 62.45 & 73.15 & 61.17 & 71.30 & 73.71 & 67.59 & 68.17 & 63.40 & 75.20 & 72.06 & 66.55 & 70.53 & 62.13 & 77.65 & 68.06 \\ 
EATA + TTAC & 62.28 & 65.54 & 65.59 & 71.90 & 59.06 & 69.63 & 72.13 & 66.00 & 66.47 & 63.38 & 72.97 & 69.55 & 63.85 & 69.06 & 60.82 & 75.21 & 67.09 \\
EATA + \textbf{Ours} & 61.29 & 65.66 & 65.32 & 74.31 & 62.79 & 72.41 & 74.77 & 69.16 & 69.95 & 65.99 & 76.22 & 73.76 & 67.75 & 71.78 & 63.42 & 77.99 & \textbf{69.53} \\ \hline
CoTTA & 59.53 & 62.34 & 60.73 & 72.02 & 62.37 & 70.48 & 72.09 & 65.86 & 66.73 & 59.08 & 72.97 & 69.69 & 65.16 & 69.20 & 63.89 & 74.28 & 66.65 \\
CoTTA + \textbf{Ours} & 60.22 & 63.06 & 62.35 & 73.23 & 62.37 & 71.40 & 73.85 & 68.84 & 68.51 & 61.79 & 75.03 & 71.93 & 66.07 & 70.68 & 63.43 & 76.62 & \textbf{68.09} \\ \hline
RMT & 62.70 & 65.69 & 64.74 & 74.54 & 67.16 & 73.98 & 76.05 & 72.87 & 73.40 & 69.66 & 77.42 & 76.11 & 74.24 & 76.23 & 71.79 & 78.25 & 72.18 \\
RMT + \textbf{Ours} & 63.21 & 67.33 & 66.86 & 74.81 & 68.47 & 74.30 & 76.11 & 73.56 & 74.07 & 70.87 & 76.94 & 76.42 & 74.79 & 76.47 & 72.93 & 77.58 & \textbf{72.79} \\ \hline
\end{tabular}
}
\end{table}

\textbf{Comparison of performance on CTA benchmarks}.
We show the effectiveness of our method in two ways, by integrating it into existing methods, and by employing the proposed loss terms independently without $\mathcal{L}_{unsup}$ (referred to as \textbf{Ours}-Only). Specifically, we apply our proposed terms on four different methods, TENT, EATA, CoTTA, and RMT, which have demonstrated promising performance on the two CTA benchmarks. \textbf{Ours} in Tab.~\ref{tab:IN} and \ref{tab:cifar100} refers to using our proposed terms $\mathcal{L}_{ema}$ and $\mathcal{L}_{src}$ together.
As illustrated in the tables, our proposed method shows noteworthy performance when used solely without $\mathcal{L}_{unsup}$ and also significantly improves performance when incorporated into existing methods. 
We also assess the performance of our method in comparison to TTAC~\citep{su2022revisiting} and TSD~\citep{wang2023feature} which are not originally designed for CTA but have been included as baseline algorithms because their proposed ideas closely align with the philosophy underlying our approach. 
TTAC tries to align the distributions between the source and target by minimizing the Kullback-Leibler (KL) divergence, under the assumption that both domains follow a Gaussian distribution. While TTAC shares a similar motivation with our $\mathcal{L}_{src}$, our approach is much simpler and more efficient. 
TSD also introduces a concept akin to our $\mathcal{L}_{ema}$, but there is a fundamental difference in that TSD utilizes a memory bank to store past test inputs, whereas our method maintains class-wise target prototypes via EMA which is more memory efficient. 
As demonstrated in the table, our proposed method consistently outperforms them despite the similarity of ideas, in the two benchmarks. We have also evaluated performance of TTAC applied to EATA, EATA+TTAC. Its performance on ImageNet-C is comparable to EATA+\textbf{Ours}, but it falls slightly short. Moreover, TTAC exhibits significant fluctuation in adaptation time depending on the target domain which will be further studied in the next section.

\begin{figure}[t]
\centering
\includegraphics[width=0.85\linewidth]{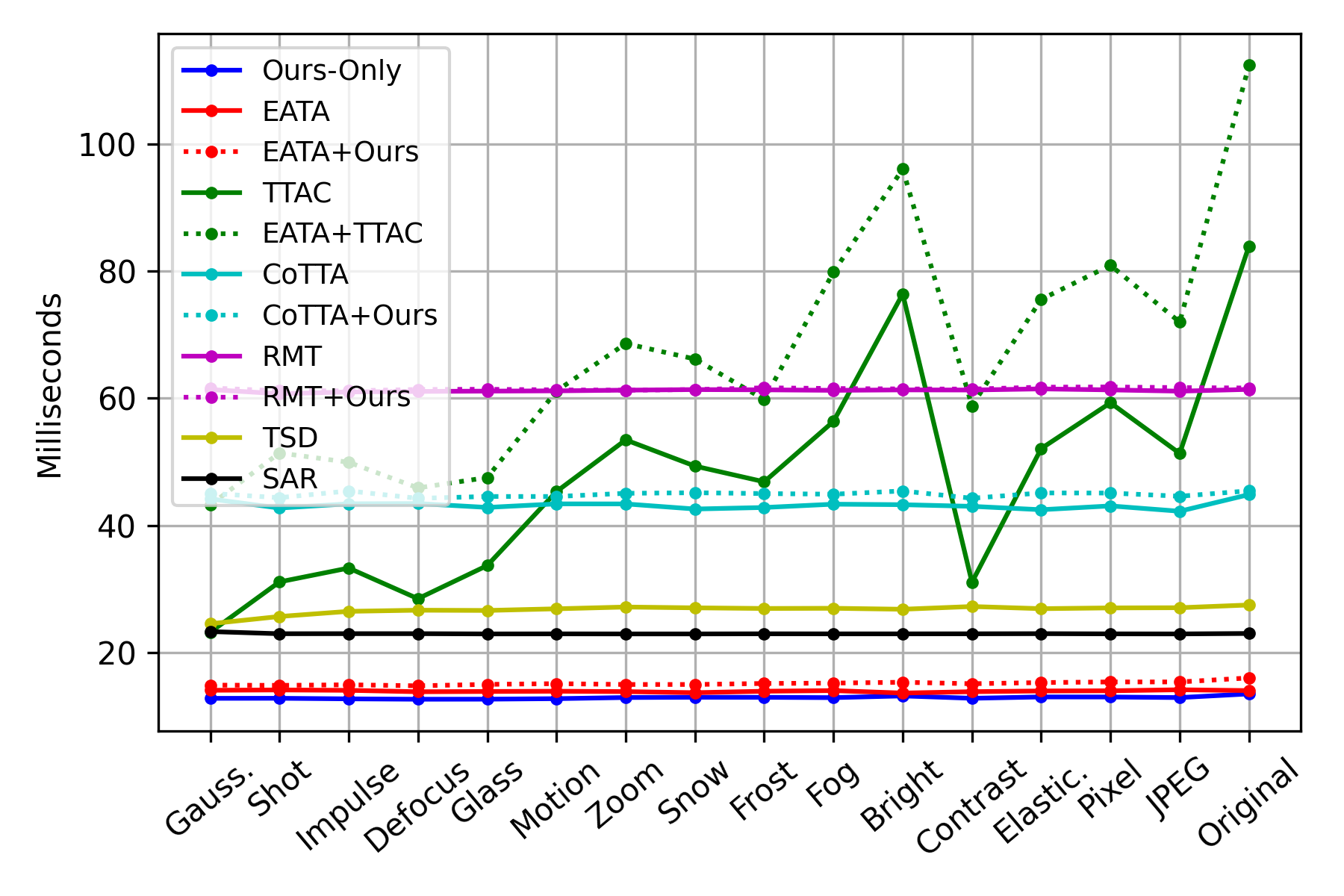} 
\caption{Comparison of average adaptation time of a single batch across target domains on ImageNet-C.}
\label{fig:adaptation_time}
\end{figure}

\textbf{Adaptation time comparison}. Adaptation time is an important factor to consider in CTA, where the model has to predict and adapt immediately in an online manner. Therefore, we measure the average time it takes to adapt a batch for each target domain and compare between methods. The experiment is conducted on a single NVIDIA RTX 3090 GPU with a fixed batch size of 64 for fair comparison. Fig.~\ref{fig:adaptation_time} illustrates the comparison of the average adaptation time of a single batch between methods across target domains of ImageNet-C. 
What stands out is the results of TTAC. Its average adaptation time of a batch exhibits significant variability across the target domains. This is attributed to TTAC's calculation of the covariance matrix using only samples with high confidence. It implies that more computational effort is needed for a particular domain which the model predicts with high confidence.
On the other hand, \textbf{Ours}-Only shows not only consistent adaptation time across the target domains but also the least amount of time required. Even when applied on existing methods such as EATA, CoTTA, and RMT, it incurs only a marginal adaptation time overhead. From the results of Tab.~\ref{tab:IN} and Fig.~\ref{fig:adaptation_time}, we demonstrate that our proposed method is able to improve the accuracy only with a negligible amount of adaptation time overhead.

\begin{table}[t]
\centering
\caption{Results of random order of ImageNet-C target domains.}\label{tab:rand_seq}
\scalebox{0.9}{
\tabcolsep4pt
\begin{tabular}{l|l|l|c} \hline
Method & Acc. (\%) & Method & Acc. (\%) \\ \hline
TTAC & 41.22$\pm$0.72 & EATA+TTAC & 50.68$\pm$0.22 \\ \hline
SAR & 41.25$\pm$1.13 & \textbf{Ours}-Only & 45.98$\pm$0.24 \\ \hline
TENT & 14.50$\pm$1.43 & TENT+\textbf{Ours} & 44.61$\pm$0.24  \\ \hline
EATA & 49.56$\pm$0.28 & EATA+\textbf{Ours} & 50.91$\pm$0.23  \\ \hline
CoTTA & 37.73$\pm$0.09 & CoTTA+\textbf{Ours} & 46.78$\pm$0.17  \\ \hline
RMT & 44.72$\pm$0.58 & RMT+\textbf{Ours} & 45.11$\pm$0.61  \\ \hline
\end{tabular}
}
\end{table}

\textbf{Robustness to random order of target domains}.
Since CTA involves adapting instantly upon the arrival of the target inputs as they arrive sequentially, the order in which the domains are presented can significantly impact the model's performance. The original domain sequence consists of consecutive domains within the same categories (noise, blur, weather, digital), making it easier to gradually adapt.
In contrast to the original sequence, we randomly shuffle the order of the 15 corrupted target domains of ImageNet-C and place the original source domain at the end. This randomization allows us to evaluate the robustness of each method to the presentation order of the target domains.
We compute the average accuracy over the 16 domains based on three separate runs, each with a distinct domain order. As shown in Table~\ref{tab:rand_seq}, the results reveal that certain methods exhibit improved performance, while others experience a decrease in performance compared to the original domain sequence. 
Notably, \textbf{Ours}-Only and methods enhanced with our approach demonstrate increased resilience to variations in the order of domains, consistently achieving superior performance when compared to the baseline methods.

\begin{table}[t]
\centering
\caption{Ablation study of proposed components on ImageNet-C.}\label{tab:IN_ablation}
\scalebox{0.85}{
\begin{tabular}{lllll|c}\hline
EATA & $\mathcal{L}_{ema}$ & $\mathcal{L}_{src}$ & Normal. & Filter. & Mean \\ \hline
\Checkmark & $-$ & $-$ & $-$ & $-$ & 49.81 \\ \hline
\Checkmark & \Checkmark & $-$ & \Checkmark & \Checkmark & 50.56 \\ 
\Checkmark & $-$ & \Checkmark & $-$ & \Checkmark & 50.80 \\ \hline
\Checkmark & \Checkmark & \Checkmark & $-$ & $-$ & 50.68\\ 
\Checkmark & \Checkmark & \Checkmark & $-$ & \Checkmark & 50.95\\ 
\Checkmark & \Checkmark & \Checkmark & \Checkmark & $-$ & 51.11\\ \hline
\Checkmark & \Checkmark & \Checkmark & \Checkmark & \Checkmark & 51.32 \\ \hline
\end{tabular}
}
\end{table}

\subsection{Analysis}
\label{sec:hyperparam_analysis}
In the following analysis, all experiments are conducted on ImageNet-C with ResNet50.

\textbf{Ablation study}.
In Tab.~\ref{tab:IN_ablation}, we assess the validity of each component of our proposed method by gradually incorporating them into the baseline algorithm, EATA. We report the mean accuracy over the 16 test domains.
The term `Normal.' in the table refers to normalizing $\omega$ and $f_{\phi}(\tilde{x}^t)$ when initializing and updating $P^t$, while `Filter.' indicates filtering the unreliable samples with high entropy. The second and third rows show the validation of our proposed loss terms, as performance improves when each loss term is added. Subsequently, the fourth to sixth rows demonstrate the significance of normalization and reliable sample selection.
When both techniques are not used (row 4), there is a significant performance drop compared to the full model (the last row).
The importance of normalization becomes evident as its removal leads to a significant drop in performance (row 5). While filtering also contributes to performance gains, its removal results in a minor performance drop (row 6) highlighting that our proposed method can robustly work even with unreliable samples possessing high entropy. The model shows the highest accuracy when every component is employed (last row).
Overall, the ablation study confirms the effectiveness of our proposed loss terms and specific implementations to the performance improvements.

\begin{figure}[t]
    \centering
    \includegraphics[width = 0.95\linewidth]{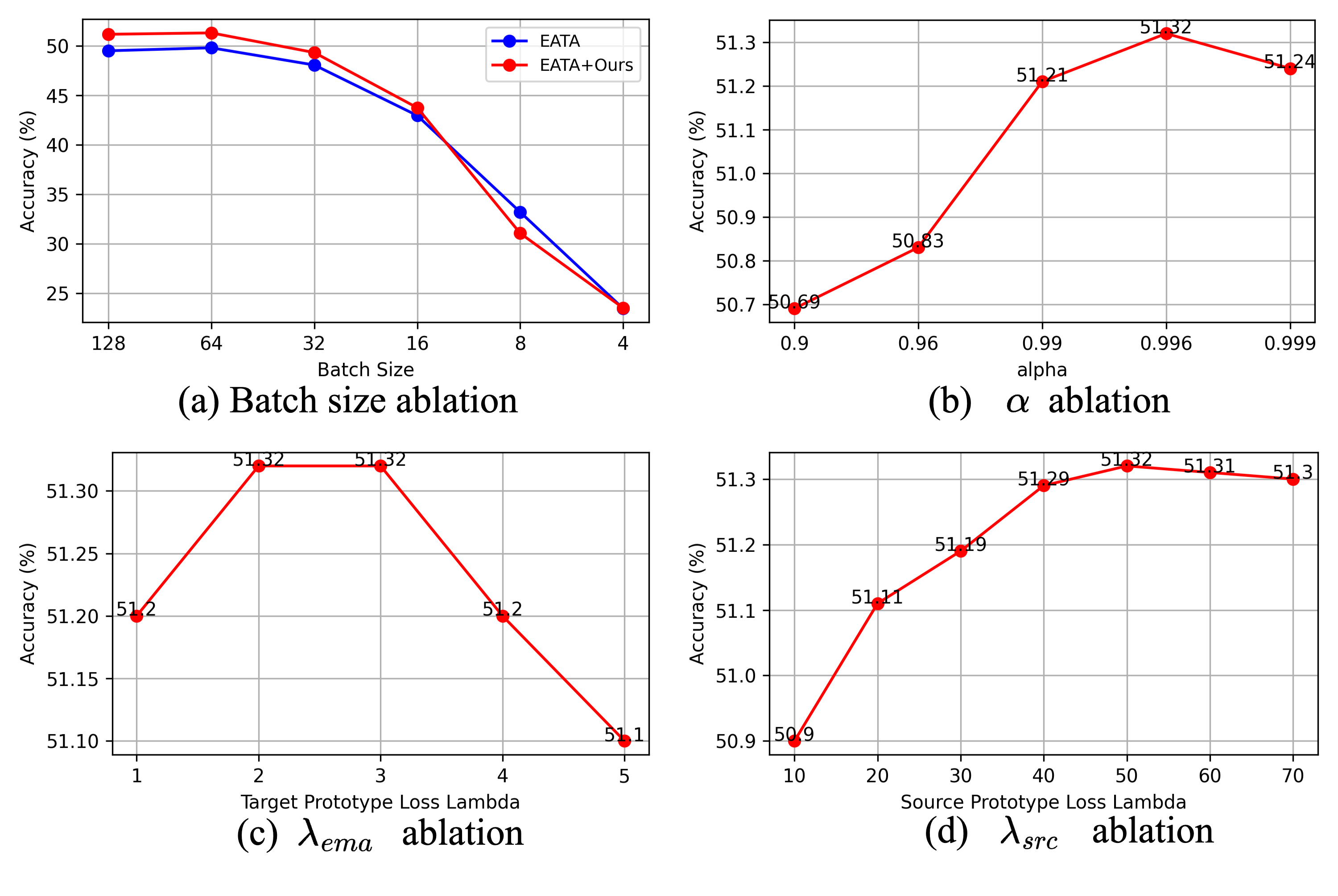}
    \caption{Analysis of batch size, $\alpha$, $\lambda_{ema}$ and $\lambda_{src}$ on ImageNet-C. (a) presents a comparison between EATA and EATA+\textbf{Ours} with varying batch sizes, while (b), (c), and (d) show performance analysis using different $\alpha$, $\lambda_{ema}$ and $\lambda_{src}$ employed in our method. Accuracy (\%) is the average accuracy over the 16 test domains.}
    \label{fig:hyperparam_analysis}
\end{figure}

\begin{figure*}[t]
    \centering
    \includegraphics[width = 0.99\linewidth]{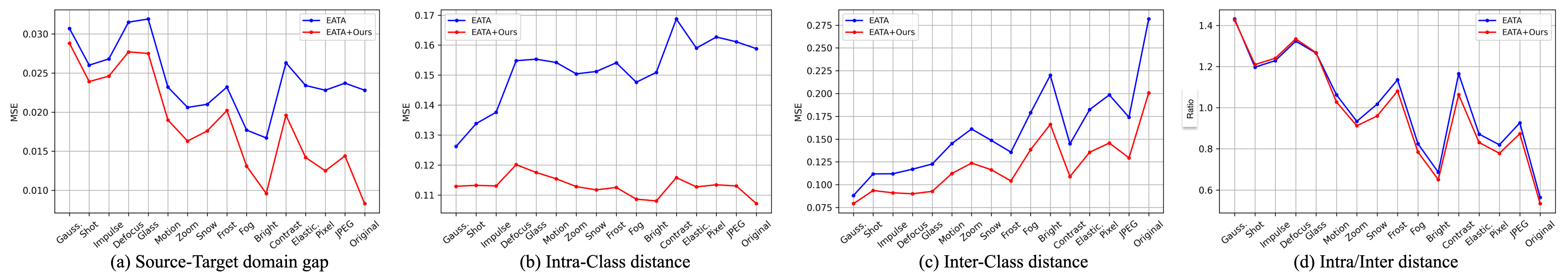}
    \caption{Feature space distance analysis. (a) plots the domain gap between the source and the target. (b) and (c) show the intra-class and the inter-class distance, respectively, while (d) presents the ratio (intra/inter) of the two distance.}
    \label{fig:feat_analysis}
\end{figure*}

\begin{figure*}[t]
    \centering
    \includegraphics[width = 0.99\linewidth]{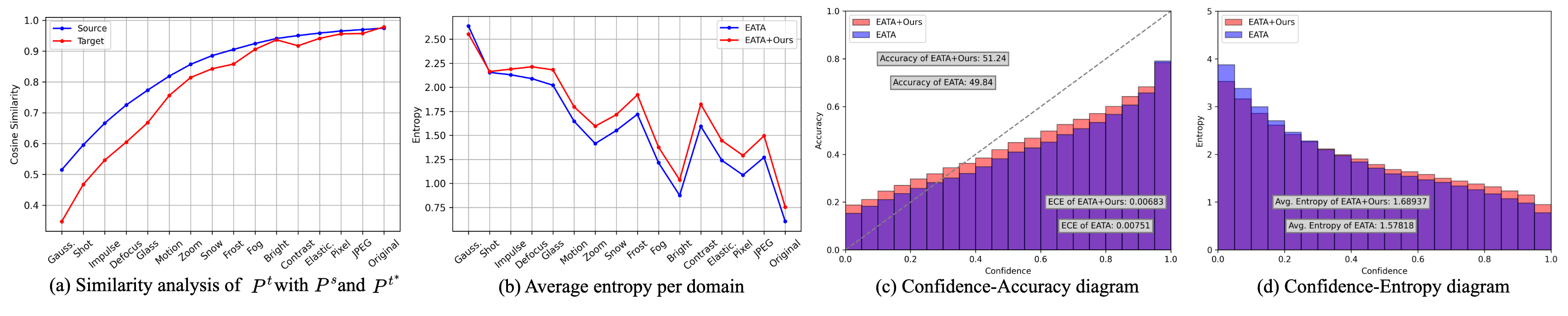}
    \caption{(a) shows the similarity analysis of $P^t$ with $P^s$ and $P^{t^*}$. (b), (c) and (d) illustrate the average entropy per domain, confidence-accuracy diagram, confidence-entropy diagram between EATA and EATA+\textbf{Ours}, respectively.
    }
    \label{fig:logit_analysis}
\end{figure*}

\textbf{Batch size}.
While it is a well-established fact that larger batch sizes often result in better model performance, the TTA setting can not guarantee large batch size as it operates online and requires immediate prediction and adaptation. Therefore, we conduct a performance comparison between EATA and EATA+\textbf{Ours} across six different batch sizes (128, 64, 32, 16, 8, 4) to evaluate the robustness of our proposed method to batch size variations. As presented in Fig.~\ref{fig:hyperparam_analysis} (a), it is evident that EATA+\textbf{Ours} consistently outperforms EATA from batch size 128 to 16. However, from a batch size of 8, both methods yield poor performance due to an extremely limited number of inputs.

\textbf{Blending factor $\alpha$}. The blending factor $\alpha$ governs the extent to which the target prototypes, $P^t$, are updated by the incoming target features. A smaller $\alpha$ promotes quicker update to new features, while a larger $\alpha$ results in a more gradual update of $P^t$, preserving the similarity to their initial states. In Fig.~\ref{fig:hyperparam_analysis} (b), we conduct an analysis of how the performance varies in EATA+\textbf{Ours} with different values of $\alpha$ (0.9, 0.96, 0.99, 0.996, 0.999). It is evident that for all five values, EATA+\textbf{Ours} outperforms the baseline algorithm EATA (49.81\%). The results clearly indicate high accuracy with large values of $\alpha$ and low accuracy with small values of $\alpha$. This observation implies that excessive update of $P^t$ with small $\alpha$ can negatively impact the model performance.

\textbf{Trade-off terms $\lambda_{ema}$ and $\lambda_{src}$}.  Fig.~\ref{fig:hyperparam_analysis} (c) and (d) provide analysis of the trade-off terms, $\lambda_{ema}$ and $\lambda_{src}$ associated with our proposed loss components, $\mathcal{L}_{ema}$ and $\mathcal{L}_{src}$ within the EATA+\textbf{Ours} model. When we vary the values of $\lambda_{ema}$, $\lambda_{src}$ is held constant at 50. Conversely, when analyzing $\lambda_{src}$, $\lambda_{ema}$ is set at 2. The model achieves its highest accuracy when $\lambda_{ema}$ is set to 2, with a decline in performance as $\lambda_{ema}$ increases. On the other hand, accuracy shows a gradual increase with rising $\lambda_{src}$ values, peaking at 50. Beyond this value, accuracy does not exhibit significant changes. Although there are differences in accuracy for various values of $\lambda_{ema}$ and $\lambda_{src}$, the gap between the highest and the lowest accuracy is relatively small. This suggests that our proposed loss terms are not highly sensitive to the choice of trade-off values.

\textbf{Source-Target distribution gap}. We analyze the distribution gap between the source and the target by measuring the MSE distance between the source prototypes and the target prototypes computed with the ground-truth (GT) labels. Unlike $P^t$ which is generated with the pseudo-labels, $P^{t^*}$ is computed with the GT labels, therefore represents the true centroid of each class cluster. During test-time adaptation, we store the features produced by $f_{\phi}$ and compute $P^{t^*}$ for each class, $P^{t^*}_c = \frac{1}{N^t_c}\sum_{i=1}^{N^t_c} f_{\phi}(x^t_i)$ where $N^t_c$ is the number of samples with GT label $c$. For each test domain, we compute the average MSE between $P^s$ and $P^{t^*}$ over the classes, $\frac{1}{C}\sum_{c=1}^{C}||P^s_c - P^{t^*}_c||^2_2$. Fig.~\ref{fig:feat_analysis} (a) illustrates the distribution gap of EATA and EATA+\textbf{Ours}. The notably lower distance observed in EATA+\textbf{Ours} compared to EATA across all test domains indicates that our proposed terms contribute significantly to narrowing the distribution gap between the source and the target domains.

\textbf{Intra- and inter-class distance of target features}. We analyze the intra-class and inter-class distance to validate how proposed method affects class-wise feature distributions. 
Intra-class distance is the average distance between the feature of every input to its corresponding $P^{t^*}$, $d^{intra}_c = \frac{1}{N^t_c}\sum_{i=1}^{N^t_c} ||P^{t^*}_c-f_{\phi}(x^t_i)||^2_2$, which can be used to validate how well the features are clustered. A smaller intra-class distance indicates that features are more effectively clustered.
Inter-class distance is the average distance between $P^{t^*}$s of different classes, which is to justify how well the clusters are separated, $d^{inter}_c = \frac{1}{C-1}\sum_{i=1}^{C} \mathbbm{1}_{\{i \neq c\}}||P^{t^*}_c-P^{t^*}_i||^2_2$. We measure both distances for each class and report the average over classes for each target domain.
In Fig.~\ref{fig:feat_analysis} (b) and (c), we present a comparison between EATA and EATA+\textbf{Ours} for both intra-class and inter-class distances. The intra-class distance of EATA+\textbf{Ours} remains consistently lower, whereas for EATA, it gradually increases, leading to a widened gap between the two methods as the adaptation progresses. It implies that our proposed terms contribute in minimizing intra-class variance.
On the other hand, concerning inter-class distance, EATA exhibits larger distances than EATA+\textbf{Ours}, suggesting that the class centroids are more widely dispersed. Nonetheless, it is noteworthy that the gap between the two methods remains relatively constant throughout the target domains when compared to the intra-class distances.
It may be tempting to conclude that EATA achieves a more class-discriminative feature distribution due to its higher inter-class distance. However, when we examine the ratio between the two distances ($d^{intra}_c/d^{inter}_c$) in Fig.~\ref{fig:feat_analysis} (d), EATA+\textbf{Ours} consistently yields lower values, especially for later target domains. A lower ratio implies a relatively larger inter-class distance compared to the intra-class distance, indicating higher class separability.

\textbf{Similarity analysis of $P^t$ with $P^s$ and $P^{t^*}$}. 
$P^t$ plays a crucial role in computing $\mathcal{L}_{ema}$. Its significance lies in its ability to accurately represent the true centroid of the class cluster. To assess its representation as the centroid of the class cluster, we analyze its cosine similarity with the prototype of the source and the target domain ($P^s$ and $P^{t^*}$) which are constructed with the ground-truth labels, hence the true centroid of the class cluster. As shown in Fig.~\ref{fig:logit_analysis} (a), it is observed that as the test-time adaptation proceeds, $P^t$ gradually shows higher similarity with both $P^s$ and $P^{t^*}$. The high similarity suggests that the EMA target prototypes, $P^t$, accurately represents the actual centroid of the class clusters. Further discussion about it continues in the appendix~\ref{app:similarity_analysis}.

\textbf{Entropy and confidence analysis}. 
Fig.~\ref{fig:logit_analysis} (b) compares an average entropy over all samples of each target domain between EATA and EATA+\textbf{Ours}.
We find an intriguing observation that the entropy of EATA+\textbf{Ours} is higher than EATA despite its superior accuracy over EATA.
This seems counterintuitive, as entropy minimization loss is widely employed for test-time adaptation. To investigate this phenomenon, we analyze the accuracy and entropy according to prediction confidence. We divide the predictions into 20 equally spaced bins based on confidence and measure the accuracy and entropy of each bin in Fig.~\ref{fig:logit_analysis} (c) and (d).
In Fig.~\ref{fig:logit_analysis} (c), the model is well calibrated when the confidence aligns with the accuracy (when the accuracy of each bin is well aligned with the grey dashed diagonal line in the figure). As depicted in the figure, EATA+\textbf{Ours} appears to be relatively more well-calibrated, exhibiting a better alignment with the dashed line. To quantitatively estimate how well the model is calibrated, we also calculate Expected Calibration Error (ECE)~\citep{naeini2015obtaining} of both models. We observe that EATA+\textbf{Ours} presents lower ECE than EATA and achieves higher accuracy in all bins except the last bin with confidence higher than 95\%. 
In Fig.~\ref{fig:logit_analysis} (d), we see that EATA+\textbf{Ours} presents lower entropy in the low confidence bins and higher entropy in high confidence bins compared to EATA. Also, as already observed in Fig.~\ref{fig:bias_in_model}, proposed method alleviates the bias in the model of favoring certain classes more and predicting with overly high confidence.
Overall, proposed method alleviates the over-confident predictions, inducing decrease in high confidence predictions and increase in low-confidence predictions. It also resolves the mis-calibration of the model which results in lower ECE. Lastly, we observe that the model achieves higher accuracy when it demonstrates low entropy on low confidence predictions and high entropy on high confidence predictions. 
We conjecture that the proposed method enhances the flexibility of model predictions by mitigating the bias, consequently aiding better generalization to target data.

\section{Conclusion}
This paper proposes a method of resolving bias in the model by exploiting prototypes of the source and the target domains for continual test-time adaptation. Its compatibility with existing methods makes it a simple yet effective plug-and-play component.
The source prototypes are employed to minimize the distribution gap between the source and the target data while the target prototypes prevent the model from being ovetfitted to the incoming target data and encourage it to capture more general distribution that can handle the changing target distribution.
Our findings reveal that it significantly improves the performance of the model with minimal adaptation time overhead. Moreover, it alleviates the bias in the model by making the model to predict less confident and to restrain from favoring certain classes more.

\section{Social Impacts}

This paper presents work whose goal is to mitigate the bias in the model for continuous test time adaptation. It can be applied to practical settings of deep learning deployment and real time adaptation. There are many potential societal consequences of our work, none which we feel must be specifically highlighted here.

\nocite{yi2023source, choi2022swr, mirza2023actmad, ganin2015dann, zhu2020cycada, tang2020discda, xu2020mixup, chen2019progressive, shen2018wdan, zhu2020cycada, style2020luo, kim2019selftraining,cst2021liu,cbst2018zou,liang2020shot,zhang2022towards, gao2023back, zhao2023delta, park2023label, prabhudesai2023diffusion, yuan2023tea, guo2017calibration}

\bibliography{main}
\bibliographystyle{icml2024}

\newpage
\appendix
\onecolumn

\section{Implementation details}\label{app:imp_details}

Here, we describe the implementation details of each method in our experiments. We use the code implemented in MECTA~\cite{hong2022mecta}\footnote{\url{https://github.com/SonyResearch/MECTA}} for TENT~\citep{wang2020tent}, EATA~\cite{niu2022eata}, and CoTTA~\citep{wang2022continual}. For other methods, we referenced official implementation of each method. We use PyTorch~\cite{NEURIPS2019_9015} framework and a single NVIDIA RTX 3090 GPU for conducting experiments.

\textbf{Tent}.~\citep{wang2020tent} We use the SGD optimizer with a learning rate of 0.0001 and a momentum of 0.9 for both ImageNet-C and CIFAR100-C datasets.

\textbf{T3A}.~\citep{iwasawa2021t3a} We referenced the official code of T3A \footnote{\url{https://github.com/matsuolab/T3A}} for its implementation. Since it is an optimization free method, there is no need for an optimizer as well as a learning rate. We use 100 for the hyper-parameter $M$ which indicates the $M$-th largest entropy of the support set.

\textbf{TSD}.~\citep{wang2023feature} We referenced the official code of TSD \footnote{\url{https://github.com/SakurajimaMaiii/TSD}} for its implementation. We use the ADAM~\citep{kingma2014adam} optimizer with a learning rate of 0.00005 for both ImageNet-C and CIFAR100-C datasets as mentioned in its paper. We use 3 for the number of nearest neighbors $K$, 100 for the entropy filter hyper-parameter $M$ and 0.1 for the trade-off parameter $\lambda$ following its implementation details described it its paper.

\textbf{TTAC}.~\citep{su2022revisiting} We referenced the official code of TTAC \footnote{\url{https://github.com/Gorilla-Lab-SCUT/TTAC}} for its implementation. 
We used the implementation version that does not use the queue since saving target data in queue at test-time costs memory and computation overhead which are not suitable for continual test-time adaptation. 
We use the SGD optimizer with a learning rate of 0.0002/0.00001 and momentum of 0.9 for ImageNet-C and CIFAR100-C datasets, respectively.
However, when we apply TTAC on EATA, we follow the implementation details of EATA and use a learning rate of 0.00025 and update only the batch normalization layers. 
We use 0.9, 0.9, 1280, 64 for $\tau_{PP}$, $\xi$, $N_{clip}$, $N_{clip\_k}$ and 0.05/0.5 for the trade-off parameter of global feature alignment, $\lambda$, in ImageNet-C and CIFAR100-C datasets, respectively, following its official implementation.

\textbf{EATA}.~\citep{niu2022eata} We use the SGD optimizer with a learning rate of 0.00025 and a momentum of 0.9 for both ImageNet-C and CIFAR100-C datasets. The entropy threshold $E_0$ is set as $0.4 \times \ln{C}$ as mentioned earlier in the main paper and the threshold for redundant sample identification, $\epsilon$, is set to 0.05. The number of samples for calculating Fisher information is set to 2000 and the trade-off parameter for anti-forgetting loss, $\beta$, is set to 2000 as well for both datasets. The moving average factor to track the average model prediction of a mini-batch for redundant sample identification is set to 0.1 as mentioned in its implementation details.

\textbf{CoTTA}.~\citep{wang2022continual} We use the SGD optimizer with a learning rate of 0.0001 and a momentum of 0.9 for the ImageNet-C dataset, whereas we employ the ADAM optimizer with a learning rate of 0.001 for CIFAR100-C. The confidence threshold for deciding whether to augment the provided inputs, denoted as $p_{th}$, is configured at 0.1/0.72, while the restore probability for generating masks for stochastic restoration, represented as $p$, is established at 0.001/0.01 for the ImageNet-C and CIFAR100-C datasets, respectively. The exponential moving average momentum for the update of the teacher model is set to 0.999 in both datasets. Originally, CoTTA uses the output of the teacher model for the evaluation, but when we apply our proposed method on CoTTA we use the output of the student for the evaluation. Also, we use the same learning rate of 
0.0001 regardless of the datasets when applying our method on CoTTA.

\textbf{RMT}.~\citep{dobler2023robust} \footnote{\url{https://github.com/mariodoebler/test-time-adaptation}} We use the SGD optimizer with a learning rate of 0.01 and a momentum of 0.9 for the ImageNet-C dataset, whereas we employ the ADAM optimizer with a learning rate of 0.0001 for CIFAR100-C. The number of samples for warm up is set to 50,000 and the trade-off parameters for contrastive loss and the source replay loss are set as 1. The temperature for contrastive loss and the exponential moving average momentum for teacher model update are set to 0.1 and 0.999, respectively. Note that RMT is not a source-free method since is employs source-replay loss during test-time adaptation which requires source domain data even at the test-time. Other than the source replay loss, it also employs contrastive loss which makes the overall loss term of RMT intricate. Therefore, when we apply our proposed terms on RMT, we use different values of $\lambda_{ema}$ and $\lambda_{src}$. For ImageNet-C, we use $\lambda_{ema}=$ 0.5 and $\lambda_{src}=$ 0.01 while we use $\lambda_{ema}=$ 1.0 and $\lambda_{src}=$ 0.01 in CIFAR100-C.

\textbf{SAR}.~\citep{niu2023towards} \footnote{\url{https://github.com/mr-eggplant/SAR}} We use the SGD optimizer with a learning rate of 0.00025 and a momentum of 0.9 for both ImageNet-C and CIFAR100-C datasets.

\textbf{RoTTA}.~\citep{yuan2023robust} \footnote{\url{https://github.com/BIT-DA/RoTTA}} We use the ADAM optimizer with a learning rate of 0.001/0.0001 for CIFAR100-C and ImageNet-C respectively. For other hyper-parameters, we follow the details described in its paper.

We adhere to the hyper-parameters as detailed in the paper or the official implementation of each method. Nevertheless, for some methods, we fine-tuned the learning rate to better align with our continual test-time adaptation setting, maintaining a fixed batch size of 64.

\begin{table*}[t]
\centering
\caption{Ablation study of consistency loss on ImageNet-C using the corruption level 5.}\label{tab:consistency_loss}
\scalebox{0.6}{
\tabcolsep4pt
\begin{tabular}{l|llllllllllllllll|c}\hline
Time & \multicolumn{16}{l|}{$t\xrightarrow{\hspace*{16.5cm}}$}& \\ \hline
Method & \rotatebox[origin=c]{70}{Gauss.} & \rotatebox[origin=c]{70}{shot} & \rotatebox[origin=c]{70}{impulse} & \rotatebox[origin=c]{70}{defocus} & \rotatebox[origin=c]{70}{glass} & \rotatebox[origin=c]{70}{motion} & \rotatebox[origin=c]{70}{zoom} & \rotatebox[origin=c]{70}{snow} & \rotatebox[origin=c]{70}{frost} & \rotatebox[origin=c]{70}{fog}  & \rotatebox[origin=c]{70}{bright.} & \rotatebox[origin=c]{70}{contrast} & \rotatebox[origin=c]{70}{elastic.} & \rotatebox[origin=c]{70}{pixelate} & \rotatebox[origin=c]{70}{jpeg} & \rotatebox[origin=c]{70}{original} & Mean \\ \hline
EATA & 34.66 & 40.40 & 39.39 & 34.08 & 34.99 & 46.51 & 52.82 & 50.33 & 45.83 & 59.12 & 67.27 & 45.17 & 57.13 & 59.99 & 55.46 & 73.80 & 49.81 \\
EATA + \textbf{Ours} & 36.17 & 41.77 & 40.83 & 35.98 & 37.24 & 48.89 & 54.28 & 52.15 & 47.46 & 60.23 & 67.94 & 48.01 & 58.26 & 61.26 & 56.37 & 74.20 & 51.32 \\
EATA + \textbf{Ours} + $\mathcal{L}_{cons}$ & 36.66 &42.33 &41.41 &36.25 &37.57 &48.91 &54.04 &52.58 &47.65 &60.34 &67.94 &48.39 &58.22 &61.36 &56.56 &74.27 &\textbf{51.53}\\ 
EATA + \textbf{Ours} + $\mathcal{L}_{cons}$(CoTTA-Aug) & 35.15 &40.30 &39.50 &33.92 &35.83 &47.38 &53.06 &51.20 &46.62 &59.54 &67.26 &47.12 &57.48 &60.49 &55.77 &73.72 &50.27\\ \hline
CoTTA & 16.15 & 18.53 & 19.91 & 18.52 & 19.58 & 31.13 & 43.07 & 36.92 & 36.15 & 51.18 & 65.35 & 23.50 & 47.71 & 52.17 & 44.82 & 73.99 & 37.42 \\
CoTTA + \textbf{Ours} & 30.06 & 37.51 & 36.72 & 26.86 & 30.65 & 42.34 & 49.64 & 47.53 & 44.15 & 56.65 & 67.13 & 37.73 & 55.98 & 59.81 & 54.68 & 73.17 & 46.91 \\
CoTTA + \textbf{Ours} + $\mathcal{L}_{cons}$ & 31.38 & 39.62 &38.97 &28.78 &32.16 &43.25 &50.39 &48.93 &44.34 &57.10 &67.07 &39.35 &55.69 &59.74 &54.75 &72.49 &\textbf{47.75}\\ 
CoTTA + \textbf{Ours} + $\mathcal{L}_{cons}$(CoTTA-Aug) & 27.57 &34.75 &35.07 &27.60 &30.50 &42.37 &49.56 &46.66 &43.31 &55.85 &66.73 &39.35 &54.70 &58.77 &53.22 &73.19 &46.20\\ \hline
\end{tabular}
}
\end{table*}

\section{Consistency loss with strong augmentation}
Employing consistency loss between original input and its augmented version is a widely used technique in semi/self-supervised learning to improve the generalization capacity of the model~\citep{chen2020simple, he2020momentum, grill2020bootstrap, liu2021unbiased, sohn2020simple}. Since TTA is also a kind of unsupervised learning, it adopts such strategy as well. CoTTA~\citep{wang2022continual} is the first TTA work to propose the use of EMA teacher network and employing the consistency loss between the outputs of the teacher and the outputs of the student with various augmentations on the inputs to the teacher network. However, we find that consistency loss can achieve better performance with stronger augmentation strategy and even without the use of the teacher network.

We do not employ the teacher network and give two versions of input (original and strong augmented version) to the network. Instead of using the augmentations used in CoTTA, we adopts augmentations proposed in \cite{liu2021unbiased} which employs randomly adding color jittering, grayscale, Gaussian blur, and cutout patches.
\begin{align}\label{eq:consist_loss}
    \mathcal{L}_{cons}(g_{\theta}, x^t, \mathcal{A})=-\sum^C_c(\sigma(g_{\theta}(x^t)) \cdot \log(\sigma(g_{\theta}(\mathcal{A}(x^t)))))^c
\end{align}
The consistency loss is defined as the cross-entropy loss between the outputs of the two inputs (original and its augmented version) predicted by the same network $g_{\theta}$ where $\mathcal{A}$ and $\sigma$ refer to the augmentation and the softmax operation.
$\mathcal{L}_{cons}$ can be additionally incorporated with a balancing trade-off parameter, $\lambda_{cons}$ which makes the overall objective as follows:
\begin{align}\label{eq:consist_overall}
    \mathcal{L}_{overall}=\mathcal{L}_{unsup} + \lambda_{ema}\mathcal{L}_{ema} + \lambda_{src}\mathcal{L}_{src} + \lambda_{cons}\mathcal{L}_{cons}.
\end{align}
We apply the consistency loss to both EATA+\textbf{Ours} and CoTTA+\textbf{Ours} to demonstrate its effectiveness. Table~\ref{tab:consistency_loss} presents the respective results, clearly indicating that $\mathcal{L}_{cons}$ contributes to performance improvement. Particularly, its impact is more pronounced when applied to CoTTA. However, when we use the augmentation strategies proposed in CoTTA for $\mathcal{A}$, denoted as $\mathcal{L}_{cons}$(CoTTA-Aug) in the table, the performance rather deteriorates. This result emphasizes the importance of using a proper augmentation strategy for the consistency loss. Our experiment suggests that using strong augmentation such as random cutout patches is indeed effective.

\begin{table*}[th]
\centering
\caption{Ablation study of $\lambda_{ema}$ on \textbf{Ours}-Only using the ImageNet-C}\label{tab:lambda_ema_ours_only}
\scalebox{1.0}{
\tabcolsep4pt
\begin{tabular}{l|l|l|l|l|l}\hline
$\lambda_{ema}$ & 1 & 2 & 3 & 4 & 5 \\ \hline
Acc. (\%) & 45.20 &	\textbf{45.96} & 44.69 &	34.29 &	26.55 \\ \hline
\end{tabular}
}
\end{table*}

\begin{table*}[th]
\centering
\caption{Ablation study of $\lambda_{src}$ on \textbf{Ours}-Only using the ImageNet-C}\label{tab:lambda_src_ours_only}
\scalebox{1.0}{
\tabcolsep4pt
\begin{tabular}{l|l|l|l|l|l|l|l}\hline
$\lambda_{src}$ & 10 & 20 & 30 & 40 & 50 & 60 & 70 \\ \hline
Acc. (\%) & 45.58 &	\textbf{45.96} &	45.86 &	45.65 &	45.29 &	43.44 &	41.01 \\ \hline
\end{tabular}
}
\end{table*}

\begin{table*}[t]
\centering
\caption{Performance comparison between soft label and hard label for $\mathcal{L}_{ema}$ on ImageNet-C}\label{tab:soft_hard_label}
\scalebox{0.63}{
\tabcolsep4pt
\begin{tabular}{l|llllllllllllllll|c}\hline
Time & \multicolumn{16}{l|}{$t\xrightarrow{\hspace*{16.5cm}}$}& \\ \hline
Method & \rotatebox[origin=c]{70}{Gauss.} & \rotatebox[origin=c]{70}{shot} & \rotatebox[origin=c]{70}{impulse} & \rotatebox[origin=c]{70}{defocus} & \rotatebox[origin=c]{70}{glass} & \rotatebox[origin=c]{70}{motion} & \rotatebox[origin=c]{70}{zoom} & \rotatebox[origin=c]{70}{snow} & \rotatebox[origin=c]{70}{frost} & \rotatebox[origin=c]{70}{fog}  & \rotatebox[origin=c]{70}{bright.} & \rotatebox[origin=c]{70}{contrast} & \rotatebox[origin=c]{70}{elastic.} & \rotatebox[origin=c]{70}{pixelate} & \rotatebox[origin=c]{70}{jpeg} & \rotatebox[origin=c]{70}{original} & Mean \\ \hline
EATA & 34.66 & 40.40 & 39.39 & 34.08 & 34.99 & 46.51 & 52.82 & 50.33 & 45.83 & 59.12 & 67.27 & 45.17 & 57.13 & 59.99 & 55.46 & 73.80 & 49.81 \\
EATA + \textbf{Ours} Hard Label & 36.17 & 41.77 & 40.83 & 35.98 & 37.24 & 48.89 & 54.28 & 52.15 & 47.46 & 60.23 & 67.94 & 48.01 & 58.26 & 61.26 & 56.37 & 74.20 & 51.32 \\
EATA + \textbf{Ours} Soft Label &  35.89 &41.60 &40.80 &35.72 &37.30 &48.82 &54.33 &52.07 &47.42 &60.28 &68.04 &48.05 &58.35 &61.29 &56.34 &74.31 &51.29\\ \hline
\end{tabular}
}
\end{table*}

\begin{table*}[t]
\centering
\caption{Performance comparison between student output and teacher output of CoTTA + \textbf{Ours} on ImageNet-C}\label{tab:cotta_output}
\scalebox{0.61}{
\tabcolsep4pt
\begin{tabular}{l|llllllllllllllll|c}\hline
Time & \multicolumn{16}{l|}{$t\xrightarrow{\hspace*{16.5cm}}$}& \\ \hline
Method & \rotatebox[origin=c]{70}{Gauss.} & \rotatebox[origin=c]{70}{shot} & \rotatebox[origin=c]{70}{impulse} & \rotatebox[origin=c]{70}{defocus} & \rotatebox[origin=c]{70}{glass} & \rotatebox[origin=c]{70}{motion} & \rotatebox[origin=c]{70}{zoom} & \rotatebox[origin=c]{70}{snow} & \rotatebox[origin=c]{70}{frost} & \rotatebox[origin=c]{70}{fog}  & \rotatebox[origin=c]{70}{bright.} & \rotatebox[origin=c]{70}{contrast} & \rotatebox[origin=c]{70}{elastic.} & \rotatebox[origin=c]{70}{pixelate} & \rotatebox[origin=c]{70}{jpeg} & \rotatebox[origin=c]{70}{original} & Mean \\ \hline
CoTTA & 16.15 &18.53 &19.91 &18.52 &19.58 &31.13 &43.07 &36.92 &36.15 &51.18 &65.35 &23.50 &47.71 &52.17 &44.82 &73.99 &37.42 \\
CoTTA + \textbf{Ours} Teacher Output & 21.75 &33.04 &36.38 &25.07 &30.68 &39.15 &47.03 &41.41 &41.80 &52.41 &65.50 &35.47 &51.56 &56.23 &51.52 &72.38 &43.84\\ 
CoTTA + \textbf{Ours} Student Output & 30.06 &37.51 &36.72 &26.86 &30.65 &42.34 &49.64 &47.53 &44.15 &56.65 &67.13 &37.73 &55.98 &59.81 &54.68 &73.17 &46.91 \\ \hline
\end{tabular}
}
\end{table*}

\section{Ablation study on trade-off terms $\lambda_{ema}$ and $\lambda_{src}$ of \textbf{Ours}-Only}
As mentioned in the implementation details described in Section~\ref{sec:exp}, when our proposed loss terms are used independently without integration into existing methods, we use $\lambda_{ema}$=2 and $\lambda_{src}$=20. 
Table~\ref{tab:lambda_ema_ours_only} and ~\ref{tab:lambda_src_ours_only} show the ablation study of $\lambda_{ema}$ and $\lambda_{src}$ with different values when our proposed terms are solely used without $\mathcal{L}_{unsup}$. When examining the effect of $\lambda_{ema}$, $\lambda_{src}$ is set at 20, whereas when investigating the impact of $\lambda_{src}$, $\lambda_{ema}$ is configured to 2. The accuracy in the tables are an average accuracy over the 16 test domains. Table~\ref{tab:lambda_ema_ours_only} illustrates that the performance reaches its peak at $\lambda_{ema}=2$, and it experiences a sharp decline when value exceeds 3. Similarly, Table~\ref{tab:lambda_src_ours_only} reveals that similar performance is maintained from 10 to 50, achieving over 45\% accuracy, but it sharply declines when value surpasses 50.

\section{Comparison of hard label and soft label for $\mathcal{L}_{ema}$}
We use the pseudo-label $\tilde{y}^t$ when calculating $\mathcal{L}_{ema}$. The pseudo-label can take the form of a one-hot vector, serving as a hard label, or it can be used as the raw logit output of the model, acting as a soft label. 
When using the soft-label, we minimize the cross-entropy loss between the output of the EMA target prototypes and the soft pseudo-label. The output of the EMA target prototypes refers to a logit, $z^t_{ema} \in \mathbb{R}^C$, produced by dot-producting $f_{\phi}(x^t)$ with every $P^t_c$ for each class.
In the main paper, we present results using the hard label representation. However, to delve deeper into the mechanism of $\mathcal{L}_{ema}$, we conduct a performance comparison using both versions of the pseudo-label, as summarized in Table~\ref{tab:soft_hard_label}. As demonstrated in the table, there is no significant distinction between the two versions of the pseudo-label, although the hard-label version exhibits slightly better performance.

\section{Comparison of student output and teacher output of CoTTA+\textbf{Ours}}
As specified in the implementation details, CoTTA originally uses the output of the teacher network for evaluation, but we employ the output of the student network when applying our proposed loss terms on CoTTA. Table~\ref{tab:cotta_output} presents a performance comparison between CoTTA+\textbf{Ours} using the output of the teacher and the output of the student. As demonstrated in the table, using the teacher network's output yields inferior performance compared to the student network's output, yet it still significantly outperforms CoTTA. We hypothesize that the reason for the student output's superior accuracy is that our proposed loss terms directly impact the student network, whereas the teacher network undergoes slow updates through exponential moving average.

\section{Similarity analysis of $P^t$ with $P^s$ and $P^{t^*}$.} \label{app:similarity_analysis}
\begin{figure}[t]
    \centering
    \includegraphics[width = 0.5\linewidth]{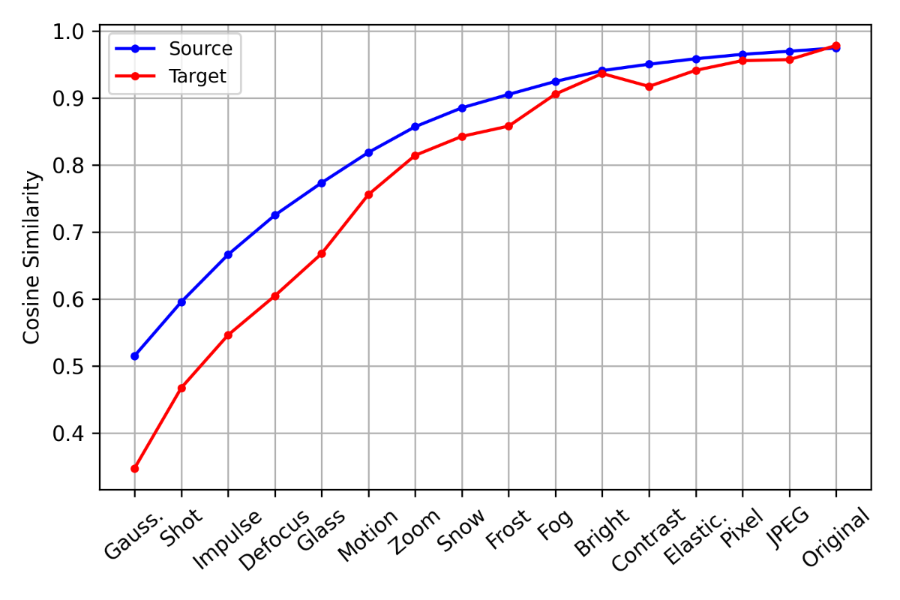}
    \caption{Cosine similarity analysis of $P^t$ with $P^s$ and $P^{t^*}$ for each target domain as the adaptation proceeds.}
    \label{fig:cos_sim_prototypes}
\end{figure}
Fig.~\ref{fig:cos_sim_prototypes} shows the results of our similarity analysis of $P^t$ with $P^s$ and $P^{t^*}$. After the model sees all the samples of a target domain, we measure the cosine similarity between the $P^t$s and the $P^s$s and the $P^t$s and $P^{t^*}$s for the target domain. We report the cosine similarity averaged over the classes, $\frac{1}{C}\sum_{c=1}^{C}cos(P^t_c, P^s_c or P^{t^*}_c )$ where $cos$ denotes cosine similarity.
The blue plot shows the similarity with the source prototypes, $P^s$, while the red plot shows the similarity with the target prototypes $P^{t^*}$. Note that $P^{t^*}$s are computed using the ground truth labels, so they represent the actual centroids of the class clusters of the target domains.
As shown in the figure, as the adaptation proceeds, the similarity with both the source and the target prototypes increase. It implies that as $P^t$s are slowly updated in an EMA manner with the features of the reliable target samples, they better represent the true centroids of the class clusters.
We also observe that the similarity with the source prototypes smoothly increases as the adaptation goes on. We conjecture this is due to our proposed source prototype alignment loss $\mathcal{L}_{src}$ which regulates the feature extractor $f_{\phi}$ to align the target feature distribution to that of the source. 
Also, the tendency of increasing similarity with the target prototypes, $P^{t^*}$ indicates that even though $P^{t}$s are updated using the pseudo-label information, since only reliable samples are employed, they succeed in maximizing similarity with the ground-truth prototypes, $P^{t^*}$. In summary, this analysis justifies the employment of our suggested EMA target prototypes.

\section{Prediction bias analysis of each target domain}\label{app:pred_bias}
\begin{figure}[t]
    \centering
    \includegraphics[width = 0.9\linewidth]{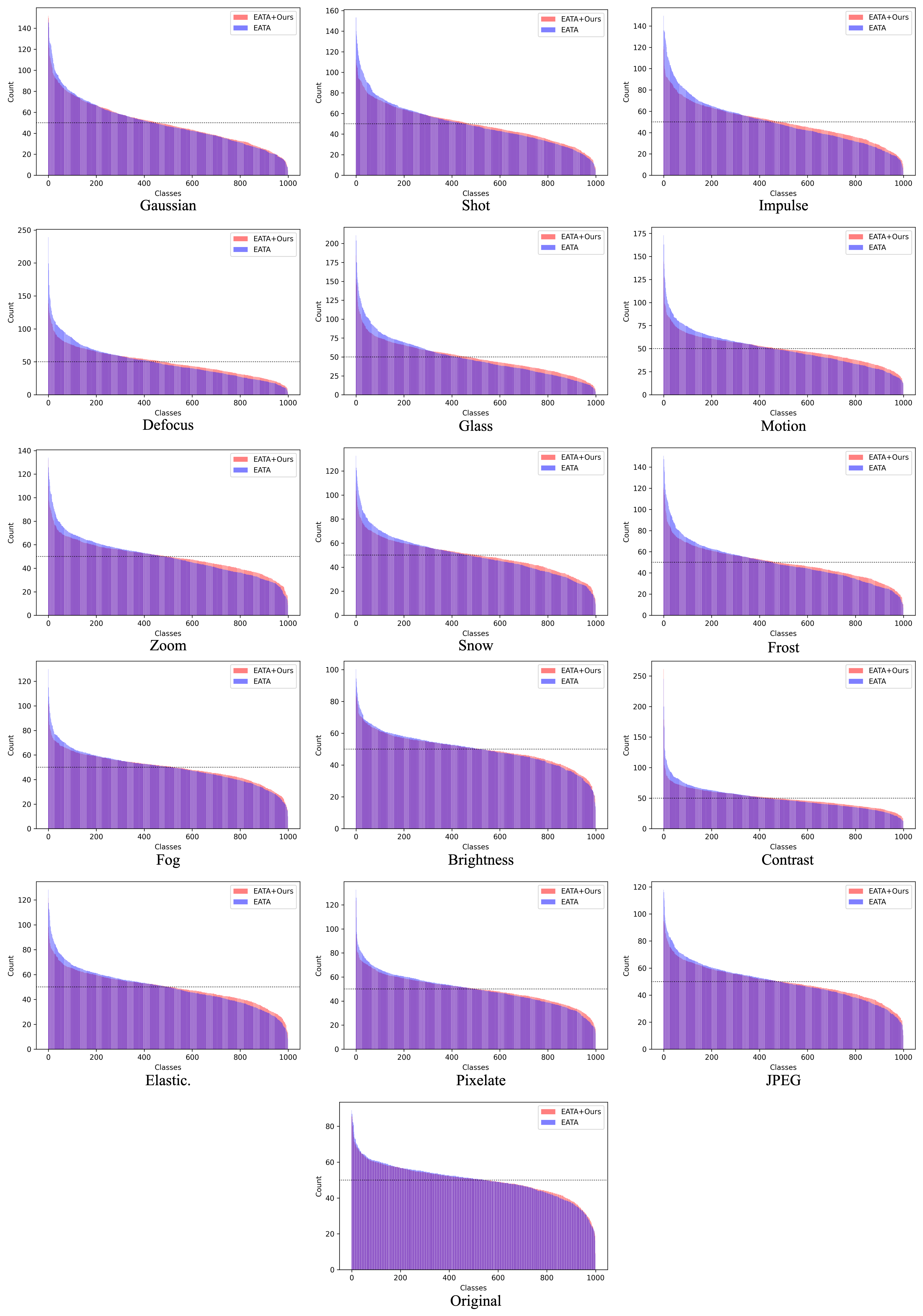}
    \caption{The comparison between EATA and EATA+\textbf{Ours} on the number of predicted samples per class for each target domain.}
    \label{fig:pred_bias}
\end{figure}
In Fig.~\ref{fig:bias_in_model} (a), we compared the number of predicted samples per class between EATA and EATA+\textbf{Ours}, demonstrating that our proposed terms contribute to a more unbiased prediction of the model, encouraging the model to predict more evenly across classes. Since Fig.~\ref{fig:bias_in_model} (a) shows the results summed over the all 16 domains, in Fig.~\ref{fig:pred_bias}, we break down the results by each domain and show the individual result of each domain. It is observed that the domains which the model shows high accuracy (brightness, original), also achieves a more balanced number of predicted samples per class across the classes. Conversely, in domains where the accuracy is low, we observe a significant bias in predictions, indicating that the model tends to favor certain classes excessively over others, making more frequent predictions on those classes. 
Overall, the bias is mitigated across all domains when our proposed terms are incorporated. EATA+\textbf{Ours} decreases predictions on the classes that EATA predicts frequently, instead, it increases predictions on the classes with a low number of predictions by EATA. Indeed, these findings confirm that our suggested terms effectively encourage the model to generate predictions that exhibit increased diversity among different classes. This mitigates the bias of the model towards favoring certain classes and, consequently, contributes to addressing the confirmation bias problem.

\section{Limitation and Future Work}
Even though our proposed EMA target prototypical loss and source distribution alignment loss indeed contribute to significant performance improvement, there are some limitations to our work that can be further developed.
The trade-off terms, $\lambda_{ema}$ and $\lambda_{src}$ for our proposed loss terms need to be fine-tuned depending on the specific method to which our proposed approach is applied.
However, we have observed that it requires minimal effort to identify suitable values for these parameters, typically falling within the range of 1 to 2 for $\mathcal{L}_{ema}$ and 20 to 50 for $\mathcal{L}_{src}$.
Also, since both $\mathcal{L}_{ema}$ and $\mathcal{L}_{src}$ rely on pseudo-labels for their computation, they can potentially result in the incorrect computation because pseudo-labels are not always accurate. To address this issue, we take measures to use only reliable samples for the computation of the loss terms. However, there is room for improvement in how we leverage pseudo-labels, such as refining them to be more precise or exploring alternative information sources for computing the loss terms.

Filtering out unreliable samples with high-entropy, is indeed an effective and efficient method to boost performance and enable efficient adaptation since it reduces the number of samples for adaptation by excluding unreliable samples. However, looking at it from a different perspective, if we can find ways to effectively harness these unreliable samples during test-time adaptation, they have the potential to make a substantial contribution to performance gains, as they represent challenging data that can introduce new insights. Disregarding high-entropy samples may inadvertently result in the loss of valuable information. Future research could focus on strategies to leverage the potential of these high-entropy samples and extract meaningful knowledge from them.
We look forward to future research endeavors that aim to tackle the aforementioned limitations and explore the suggested avenues for future work.

\end{document}